\documentclass{article}

\usepackage{arxiv}

\usepackage[utf8]{inputenc} 
\usepackage[T1]{fontenc}    
\usepackage{hyperref}       
\usepackage{url}            
\usepackage{booktabs}       
\usepackage{amsfonts}       
\usepackage{nicefrac}       
\usepackage{microtype}      
\usepackage{lipsum}

\usepackage{cite}
\usepackage{amsmath,amssymb,amsfonts}
\usepackage{graphicx}
\usepackage{textcomp}
\usepackage{xcolor}
\usepackage{xspace}
\usepackage{tikz}
\usepackage{pgfplots}
\usetikzlibrary{arrows,backgrounds,decorations,decorations.pathmorphing,positioning,fit,automata,shapes,snakes,patterns,plotmarks,calc}
\usepackage{subfig}
\usepackage[vlined,ruled,linesnumbered]{algorithm2e}
\usepackage{stmaryrd}
\usepackage[inline]{enumitem}
\usepackage{booktabs}

\newtheorem{definition}{Definition}[section]

\newcommand{\f}{\varphi}
\newcommand{\STL}{STL\xspace}

\newcommand{\PSTL}{PSTL\xspace}
\newcommand{\Intvl}{I}
\newcommand{\sPara}{c}

\newcommand{\val}{\nu}

\newcommand{\x}{\mathbf{x}}
\newcommand{\G}{\mathbf{G}}
\newcommand{\F}{\mathbf{F}}

\newcommand{\ev}{\mathbf{F}}
\newcommand{\U}{\mathbf{U}}
\newcommand{\true}{\mathit{true}}
\newcommand{\nmodels}{\not\models}
\newcommand{\aand}{\,\wedge\,}

\newcommand{\Reals}{\mathbb{R}}

\newcommand{\PosReals}{\mathbb{R}^{\ge 0}}
\newcommand{\domain}{\mathcal{D}}
\newcommand{\setof}[1]{\left\{#1\right\}}

\newcommand{\params}{\mathcal{P}}
\newcommand{\timedomain}{T}
\newcommand{\valuedomain}{V}
\newcommand{\paramspace}{{\domain_\params}}
\newcommand{\timeparams}{\params^\timedomain}
\newcommand{\valueparams}{\params^\valuedomain}
\newcommand{\p}{\mathbf{p}}
\newcommand{\Traces}{X}

\newcommand{\mypara}[1]{\vspace{0.3em} \noindent {\bf #1.\ }}

\newcommand{\db}{DB}

\newcommand{\atomIt}{\textit{atomIt}}
\newcommand{\unOpIt}{\textit{unOpIt}}
\newcommand{\binOpIt}{\textit{binaryOpIt}}
\newcommand{\unaryArgIt}{\textit{argIt}}
\newcommand{\lhsIt}{\textit{lhsIt}}
\newcommand{\rhsIt}{\textit{rhsIt}}

\newcommand{\llength}{\ell}

\newcommand{\enumatoms}{\textit{EnumAtoms}}
\newcommand{\TryClassifier}{\mathit{TryClassifier}}

\newcommand{\isNew}{\mathit{isNew}}

\title{Interpretable Classification of Time-Series Data using Efficient Enumerative Techniques}

\author{
  Sara Mohammadinejad\\
  University of Southern California\\
  \texttt{saramoha@usc.edu} \\
   \And
  Jyotirmoy V. Deshmukh\\
  University of Southern California\\
  \texttt{jdeshmuk@usc.edu} \\
   \And
  Aniruddh G. Puranic\\
  University of Southern California\\
  \texttt{puranic@usc.edu} \\
   \And
  Marcell Vazquez-Chanlatte\\
  University of California, Berkeley\\
  \texttt{marcell.vc@berkeley.edu} \\
   \And
  Alexandre Donz\'{e}\\
  Decyphir, Inc\\
  \texttt{alex@decyphir.com} \\
}

\begin{document}
\maketitle

\begin{abstract}
Cyber-physical system applications such as autonomous vehicles, wearable devices, and avionic systems generate a large volume of time-series data. Designers often look for tools to help classify and categorize the data. Traditional machine learning techniques for time-series data offer several solutions to solve these problems; however, the artifacts trained by these algorithms often lack interpretability.  On the other hand, temporal logics, such as Signal Temporal Logic (STL) have been successfully used in the formal methods community as specifications of time-series behaviors. In this work, we propose a new technique to automatically learn temporal logic formulae that are able to cluster and classify real-valued time-series data. Previous work on learning STL formulas from data either assumes a formula-template to be given by the user, or assumes some special fragment of STL that enables exploring the formula structure in a systematic fashion. In our technique, we relax these assumptions, and provide a way to systematically explore the space of all STL formulas. As the space of all STL formulas is very large, and contains many semantically equivalent formulas, we suggest a technique to heuristically prune the space of formulas considered.  Finally, we illustrate our technique on various case studies from the automotive, transportation and healthcare domain.

\end{abstract}

\section{Introduction}
Cyber-physical systems (CPS) generate large amounts of data due to a
proliferation of sensors responsible for monitoring various aspects of
the system. Designers are typically interested in extracting
high-level information from such data but due to its large volume,
manual analysis is not feasible. Hence a crucial question is: {\em How
do we automatically identify logical structure or relations within CPS
data?} Machine learning (ML) algorithms may not be specialized to
learn logical structure underlying time-series
data\cite{vazquez2017logical}, and typically require users to
hand-create features of interest in the underlying time-series
signals. These methods then try to learn discriminators over feature
spaces to cluster or classify data. These feature spaces can often be
quite large, and ML algorithms may choose a subset of these features
in an {\em ad hoc}  fashion. This results in an artifact (e.g., a
discriminator or a clustering mechanism) \cite{jones2014anomaly} that
is not human-interpretable. In fact, ML techniques focus only on
solving the classification problem and suggest no other comprehension
of the target system \cite{bombara2016decision}.

Signal Temporal Logic (STL) is a popular formalism to express
properties of time-series data in several application contexts, such
as automotive systems
\cite{kapinski2016st,jin2014powertrain,balkan2018underminer}, analog
circuits \cite{maler2013monitoring}, biology \cite{stoma2013stl},
robotics \cite{xu2015temporal}, etc. STL is a logic over Boolean and
temporal combinations of signal predicates which allows
human-interpretable specification of continuous system requirements.
For instance, in automotive domain, STL can be used to formulate
properties such as ``the car successfully stops before hitting an
obstruction'' \cite{kong2014temporal}.

There has been significant work in learning STL specifications from
data. Some of this work has focused on supervised learning (where
given labeled traces, an STL formula is learned to distinguish
positively labeled traces from negatively labeled traces), or
clustering (where signals are clustered together based on whether they
satisfy similar STL formulas), or anomaly detection (where STL is
utilized for recognizing the anomalous behavior of an embedded
system). There are two main approaches in learning STL formulas from
data: template-free learning and template-based learning.
Template-free methods learn both the structure of the underlying STL
formula and the formula itself.  While these techniques have been proven
effective for many diverse applications \cite{jones2014anomaly,
kong2014temporal,bombara2016decision}, they typically explore only a
fragment of STL, and may produce long and complicated STL classifiers.
Certain properties such as concurrent eventuality, repeated patterns
and persistence may not be expressible in the chosen fragments for
learning \cite{ kong2014temporal}.  In template-based methods, the
user provides a template parametric STL formula (PSTL formula) and the
learning algorithm infers only the values of parameters from data
\cite{vazquez2017logical,vazquez2018time,jin2015mining,
jha2017telex,jha2019telex}. Without a good understanding of the
application domain, choosing the appropriate PSTL formula can be
challenging.  Furthermore, in template-based methods the users may
provide very specific templates which may make it difficult to derive
new knowledge from data\cite{bombara2016decision}. 

Syntax-guided synthesis is a new paradigm in ``learning from
examples'', where the user provides a grammar for expressions, and the
learning algorithm tries to learn a concise expression that explains a
given set of examples.  In \cite{udupa2013transit}, systematic
enumeration has been used to generate candidate solutions. For
medium-sized benchmarks, the systematic enumeration algorithm, in
spite of its simplicity, surprisingly outperforms several other
learning approaches \cite{alur2013syntax}. 

Inspired by the idea of learning expressions from grammars, in this
paper, we consider the problem of learning STL based formulas to
classify a given labeled time-series dataset.  The key challenge in
systematic enumeration for STL is that predicates over real-valued
signals and time-bounds on temporal operators both involve real
numbers. This means that even for a fixed length, there are an
infinite number of STL formulas of that length.  One solution is to
apply the enumerative approach to PSTL, which uses parameters instead
of numbers. The inference problem then tries to learn parameter values
to separate labeled data into distinct classes. The
parameter-valuation inference procedures are typically efficient, but
over a large dataset, the cost for enumeration followed by parameter
inference can add up. As a result, we explore an optimization which
involves skipping formulas that are heuristically determined to be
equivalent.

As a concrete application of enumerative search, we consider the
problem of learning an STL-based classifier with a minimal
mis-classification rate for the given labeled dataset.

To summarize, our key contributions are as follows:
\begin{itemize}

\item We extend the work in \cite{vazquez2017logical,vazquez2018time,jin2015mining,jha2017telex,
jha2019telex} by learning the structure or template of PSTL formulas
automatically. The enumerative solver furthers the Occam's razor
principle in learning (simplest explanations are preferred). Thus, it
produces simpler STL formulas compared to existing template-free
methods \cite{jones2014anomaly,kong2014temporal,bombara2016decision}.

\item We introduce the notion of formula signature as a heuristic to
prevent enumeration of equivalent formulas. 

\item We bridge formal methods and machine learning algorithms by
employing STL, which is a language for formal specification of
continuous and discrete system behaviors \cite{maler2004monitoring}.
We use Boolean satisfaction of STL formulas as a formal measure for
measuring the mis-classification rate.

\item We showcase our technique on real world data from several
domains, including automotive, transportation and healthcare domain.

\end{itemize}

\section{Preliminaries}

\begin{definition}[Time-Series, Traces]
A trace $x$ is a mapping from time domain $\timedomain$ to value
domain $\domain$, $x: \timedomain \rightarrow \domain$ where,
$\timedomain \subseteq \PosReals$, $\domain \subseteq \Reals^n$ and $D
\neq \emptyset$.  
\end{definition}

\mypara{Signal Temporal Logic (STL)} Signal Temporal Logic
\cite{maler2004monitoring} is used as a specification language for
reasoning about properties of real-valued signals. The simplest
properties or constraints can be expressed in the form of atomic
predicates. An atomic predicate is formulated as $f(\x) \sim \sPara$,
where $f$ is a function from $\domain$ to $\Reals$, $\sim \in
\setof{\geq, \leq,=}$,  and $\sPara \in \Reals$.  For instance, $x(t)
\geq 2$ is an atomic predicate, where $f(x) = x(t)$, $\sim$ is $\geq$,
and $c=2$. Temporal properties include temporal operators such as $\G$
(always), $\F$ (eventually) and $\U$ (until). For example, $\G (x(t) >
2)$ means signal $x(t)$ is always greater than 2. Each temporal
operator is indexed by an interval $\Intvl := (a,b) \mid (a,b] \mid
[a,b) \mid [a,b]$, where $a, b \in \timedomain$. Every STL formula is
written in the following form:
\begin{equation}
\label{eq:stl_syntax}
\begin{array}{l}
\f :=      \true 
      \mid f(\x) \sim \sPara 
      \mid \neg\f \mid \f_{1} \wedge \f_{2} 
      \mid \f_{1}\, \U_{\Intvl}\, \f_{2}~
\end{array}
\end{equation}
where $c \in \Reals$. $\G$ and $\F$ operators are special instances of
$\U$ operator \mbox{$\F_{\Intvl}\varphi \triangleq \true\, \U_\Intvl\,
\varphi$}, and \mbox{$\G_{\Intvl}\varphi \triangleq \neg \F_{\Intvl}
\neg \varphi$}, and they are defined for formula simplification. The
boolean semantics of an \STL formula are defined formally as follows:
\begin{equation*}
\label{eq:stl_semantics}
\begin{array}{lcl}
  (\x,t) \models f(\x) \sim c &\iff& \text{$f(\x(t)) \sim c$ is true} \\
  (\x,t) \models \neg \f &\iff&  (\x,t) \nmodels \f \\
  (\x,t) \models \f_1 \wedge \f_2 &\iff& (\x,t)
  \models \f_1\ \aand\  (\x,t) \models \f_2\\
  (\x,t) \models \f_1\ \U_{\Intvl}\ \f_2 & \iff & \exists t_1\in t \oplus \Intvl: (\x,t_1) \models \f_2\ \aand \\
                                         &      & \quad \forall t_2\in[t, t_1): (\x,t_2) \models \f_1
\end{array}
\end{equation*}
$\x \models \f$ is the shorthand of $(\x,0) \models \f$. The signal
$x$ satisfies $f(x) > 0$ at time $t$ (where $t \geq 0)$ if $f(x(t)) >
0$. It satisfies $\f = \G_{[0,2)} (x>0)$ if for all time $0 \leq t <
2, x(t) > 0$ and satisfies $\f = \F_{[0, 1)} (x > 0)$ if exists $t$,
such that $0 \leq t < 1$ and $x(t) > 0$. The signal $x$ satisfies the
formula $\f = (x > 0) \U_{[0,2]} (x < 1)$ if there exists some time
$t_1$ where $0 \leq t_1 \leq 2$ and $x(t_1) < 1$, and for all time
$t_2 \in [0,t_1), x(t_2) > 0$. We can create higher-level STL formulas
by utilizing two or more of the operators. For instance, a signal $x$
satisfies $\f = \F_{[0,1]} \G_{[0,2]} (x(t) > 1)$ iff there exists
$t_1$ such that $0 \leq t_1 \leq 1$, and for all time $t_1 \leq t \leq
t_1+2, x(t) > 1$.

In addition to the Boolean semantics, quantitative semantics of \STL
quantify the robustness degree of satisfaction by a particular
trace\cite{fainekos2009robustness, donze2010robustness}. Intuitively,
a \STL with a large positive robustness is far from violation, and
with large negative robustness is far from satisfaction. If the
robustness is a small positive number, a very small perturbation might
make it negative and lead to the violation of the property.
Quantitative semantics of \STL are formally defined as follow:
\begin{equation*}
\label{eq:stl_q_semantics}
\begin{array}{lcl}
\rho(\mu,\x,t) = f(\x(t)) \\
\rho(\neg\phi,\x,t) = -  \rho(\phi,\x,t)  \\
\rho(\phi_1\aand\phi_2,\x,t) =\min( \rho(\phi_1,\x,t), \rho(\phi_2,\x,t))  \\
\rho(\phi_1\U_{\Intvl}\phi_2,\x,t) = \sup\limits_{t^{\prime}\in t \oplus \Intvl} 
    \min \begin{pmatrix}\rho(\phi_2,\x,t^{\prime}),\\ 
                        \inf_{t^{\prime\prime}\in [t,t^\prime) }
                            \rho(\phi_1,\x,t^{\prime\prime})
         \end{pmatrix} 

\end{array}
\end{equation*}

For example, for \STL formula $x(t) > 2$ the robustness value is $x(t)
- 2$, and for $x(t) < 2$ the robustness is $2 - x(t)$.  Property
"Always between time 0 and some unspecified time $\tau$, the signal
$x$ is less than some value $\pi$" could be expressed using Parametric
STL (PSTL) $\G_{[0,\tau]} (x < \pi)$, where the unspecified values
$\tau$ and $\pi$ are referred to as \textit{parameters}.

\mypara{Parametric Signal Temporal Logic (\PSTL)} \PSTL
\cite{asarin2011parametric} formula is an extension of \STL formula
where constants are replaced by parameters. The associated STL formula
is obtained by assigning a value to each parameter variable using a
valuation function. Let $\params$ be the set of parameters,
$\valuedomain$ represent the domain set of the parameter variables
$\valueparams$, and $\timedomain$ represent the time domain of the
parameter variables $\timeparams$. Then $\params$ is the set
containing the two disjoint sets $\valueparams$ and $\timeparams$,
where at least one of the sets is non-empty. A valuation function
$\val$ maps a parameter to a value in its domain. A vector of
parameter variables $\p$ is obtained by mapping parameter vectors $\p$
into tuples of respective values over $\valuedomain$ or $\timedomain$.
Hence, we obtain the {\em parameter space} $\paramspace \subseteq
\valuedomain^{|\valueparams|} \times \timedomain^{|\timeparams|}$.

An \STL formula is obtained by pairing a \PSTL formula with a
valuation function that assigns a value to each parameter variable.
For example, consider the \PSTL formula $\f(c, \tau) = \F_{[0,\tau]}(x
> c)$ with parameters $c$ and $\tau$. The STL formula $\G_{[0,5]} (x >
-2.3)$ which is an instance of $\f$ is obtained with the valuation
$\val = \setof{\tau \mapsto 5, c \mapsto -2.3}$.

\begin{definition}[Monotonic PSTL] A parameter $p_i$ is said to be
monotonically increasing or have positive polarity in a \PSTL formula
$\f$ if condition \eqref{eq:mono_inc} holds for all $\x$, and is said
to be monotonically decreasing or negative polarity if condition
\eqref{eq:mono_dec} holds for all $\x$, and monotonic if it is either
monotonically increasing or decreasing \cite{asarin2011parametric}.
\begin{align}
\hspace{-2em}
\val(p_i) \le \val'(p_i)
& \Rightarrow & 
\left [\x \models \f(\val(p_i)) \Rightarrow 
            \x \models \f(\val'(p_i)) \right]\label{eq:mono_inc} \\
\hspace{-2em}
\val(p_i) \geq \val'(p_i)
& \Rightarrow & 
\left[\x \models \f(\val(p_i)) \Rightarrow \x \models
\f(\val'(p_i))\right] \label{eq:mono_dec}
\end{align}
\end{definition}
For example, in the formula $\F_{[0,\tau]} (x > c)$, polarity of
$\tau$ is positive (the formula is monotonically increasing with
respect to $\tau$), and polarity of $c$ is negative (the formula is
monotonically decreasing with respect to $c$). 

%
%

\begin{definition}[Validity domain] The validity domain for a given
set of parameters $\params$ is the set of all valuations s.t. for the
given set of traces $\Traces$, each trace satisfies the \STL formula
obtained by instantiating the given \PSTL formula with the parameter
valuation.  The boundary of the validity domain is the set of
valuations where the robustness value of the given \STL formula with
respect to at least one trace in $\Traces$ is $0$.
\end{definition}
Essentially, the validity domain boundary serves as a classifier to
separate the set of traces satisfying the \STL formula from the ones
violating the formula.



\mypara{Supervised Learning/Classification} Supervised Learning is an
ML technique used for learning from labeled data set. Supervised
classification problems are either binary (only two classes are
involved) or multi-class classification (more than two classes are
included). In this paper, we explore the problem of binary
classification for time-series data and use boolean semantics of \STL
as a logical measure for misclassification rate (MCR). In general, MCR
is computes as the number of falsely classified traces divided by the
number of all traces. We then evaluate our method on real-world data,
including automotive, transportation and healthcare domain.

\section{Enumerative Learning for STL}

In this section, we introduce systematic \PSTL enumeration for
learning \PSTL formula classifiers from time-series data, which the
enumeration procedure is formalized in Algo.~\ref{alg:formula_enum}.
From a grammar-based perspective a \PSTL formula can be viewed as
atomic formulas combined with unary or binary operators.  For
instance, \PSTL formula $\G_{[0,\tau_1]} (x(t) > c_1) \aand
\F_{[0,\tau_2]} (x(t)< c_2)$ consists of binary operator  $\aand$,
unary operators $\G$ and $\F$, and atomic predicates $x(t) > c_1$ and
$x(t)< c_2$.
\begin{equation}
\begin{array}{l}
\f := atom \mid unaryOp(\f) \mid binaryOp(\f, \f) \\
unaryOp := \neg \mid \F \mid \G\\
binaryOp := \vee \mid \wedge \mid \U_I \mid \Rightarrow 
\end{array}
\end{equation}

Algo.~\ref{alg:formula_enum} is algorithm with several nested
iterations. The outermost loop iterates over the length of the
formula, and in the first iteration of the loop, we basically
enumerate formulas of length $1$, or parameterized signal predicates
using the \textit{EnumAtoms} function.  At end of the $\ell^{th}$
iteration of the algorithm, all formulas up to length $\ell$ are
stored in a database $\db$ that is an array of lists.  Each array
index corresponds to the formula length, and each of the lists stored at
location $\ell$ is the list of all formulas of length $\ell$. In each
iteration corresponding to lengths greater than $1$, the algorithm
calls the function \textit{ApplyUnaryOps}, and in all iterations for
lengths greater than $2$, the function also calls
\textit{ApplyBinaryOps}. When enumerating formulas of length $\ell$,
\textit{ApplyUnaryOps} function iteratively applies each unary
operator from the ordered list \textit{unaryOps} to all formulas of
length $\ell-1$ to get a new formula. The \textit{ApplyBinaryOps}
iteratively applies each binary operator from the ordered list
\textit{binaryOps} to a pair of formulas of lengths $a$ and $b$, where
$a,b \in [1,\ell-2]$ and $a+b = \ell-1$. We use \atomIt, \unOpIt,
\binOpIt, \unaryArgIt, \lhsIt, \rhsIt as iterators
(indices) on the lists \textit{atoms}, \textit{unaryOps},
\textit{binaryOps}, $\db(\ell-1)$, and $\textit{lhsArgs}$ and
$\textit{rhsArgs}$ respectively.

For each PSTL formula $\varphi$ generated by
Algo.~\ref{alg:logical_classification}, we apply the procedure
($\TryClassifier$). If  $\varphi$ is a good classifier (small
misclassification rate), all loops terminate and $\varphi$ is
returned. Otherwise, the procedure continues to generate new PSTL
formulas. We explain Algo.~\ref{alg:logical_classification} in
Section~\ref{sec:logic_bin_classification}.



In order to avoid applying Algorithm \ref{alg:logical_classification}
($\TryClassifier$) on equivalent \PSTL formulas,
we use the idea of \textit{formula signatures} to heuristically detect
equivalent  \PSTL formulas.  We explain this optimization in the next
section.



\begin{algorithm}
\small
\caption{Formula enumeration algorithm \label{alg:formula_enum}}
\SetKwProg{Fn}{Function}{:}{}
\KwIn{\textit{maxLength,atoms, unaryOps, binaryOps}, $\db$}
Init: $\llength \leftarrow 1$\;
\While{$\llength \leq$ maxLength}{
	\lIf{$\llength =1$}{
		\enumatoms()
	}
	\lElseIf{$\llength = 2$}{
		\textit{ApplyUnaryOps}()
	}
	\lElse{
		\textit{ApplyUnaryOps}(); \textit{ApplyBinaryOps}() 
	}
} 
\Fn{\enumatoms()}{
	\atomIt $\leftarrow 1$\;
	\While{\atomIt $\leq \lvert atoms \rvert$}{
		$\f \leftarrow get(atoms, \atomIt)$\;
		$\TryClassifier(\f)$ \tcp*{Algorithm \ref{alg:logical_classification}} 
		$add(\db, \llength, \f)$\;
		\atomIt $\leftarrow \atomIt + 1$\;
	}
	$\llength \leftarrow \llength + 1$\;
}
\vspace{0.1cm}
\Fn{\textit{ApplyUnaryOps}()}{
	\unaryArgIt $\leftarrow 1$, \unOpIt $\leftarrow 1$ \;
	\While{\unOpIt $\leq \lvert unaryOps \rvert$}{
		$op \leftarrow get(unaryOps, \unOpIt)$\;
		\While{\unaryArgIt $ < \lvert \db(\llength-1)$}{
			$unaryArg \leftarrow get(\db(\llength-1), \unaryArgIt)$\;
			$\f \leftarrow op(unaryArg)$ \;
			$\TryClassifier(\f)$\;
			$Add(\db,\llength, \f)$\;
			\unaryArgIt $\leftarrow \unaryArgIt + 1$\;
		}
	\unOpIt $\leftarrow \unOpIt + 1$ \;
	}
}
\vspace{0.1cm}
\Fn{\textit{ApplyBinOps}()}{
		\lhsIt, \rhsIt, \binOpIt $\leftarrow 1$ \;
		\While{\binOpIt $\leq \lvert \textit{binaryOps} \rvert$}{
			$op \leftarrow get(\textit{binaryOps}, \binOpIt)$\;
			\For{$i \leftarrow 1$ to $\llength-2$}{
				\While{\lhsIt $< \lvert \db(i) \rvert$}{
					\While{\rhsIt $< \lvert \db(\llength-i-1) \rvert$}{
						$lhs \leftarrow get(\db(i), \lhsIt)$ \;
						$rhs \leftarrow get(\db(\llength-i-1), \rhsIt)$\;
						$\f  \leftarrow op(lhs, rhs)$ \;
				                $\TryClassifier(\f)$\;
				                $Add(\db,\llength, \f)$\;
						$\rhsIt \leftarrow \rhsIt + 1$ \;
					}
					$\lhsIt \leftarrow \lhsIt + 1$\;
				}
			}
			\binOpIt $\leftarrow \binOpIt + 1$ \; 
		}
}
\end{algorithm}

\section{Signature-based Optimization}
\label{sec:signature}
Enumerating logical binary classification, which is implemented in
Algorithm \ref{alg:logical_classification}, for every  PSTL formula is
time-consuming. The problem with na\"{i}ve enumeration approach for
\PSTL formulas is that many equivalent \PSTL formulas are enumerated.
Hence, we are interested in detecting equivalent \PSTL formulas in
order to decrease enumeration time. However, checking equivalence of
\PSTL formulas is in general undecidable \cite{jin2015mining}. Even if
we restrict ourselves to a fragment of \PSTL, equivalence checking is
a computationally demanding task. Thus, we use signatures to avoid
enumerating logically equivalent formulas. A \textit{signature} is an
approximation of an STL formula.  Inspired by polynomial identity
testing \cite{raz2005deterministic}, we use the notion of signature to
check the equivalence of two STL formulas. Let $\Traces_n \subseteq
\Traces$ be a randomly chosen subset of $\Traces$ (the traces) of cardinality $n$.
Let $\paramspace_m = \setof{\val_1,\ldots,\val_m}$ be a finite subset
of $\paramspace$ (the parameter space). The signature $S$ of a formula $\f$ maps $\f$ to a matrix of real numbers,
defined as below: 
\[ S_\f(i,j) =  \rho(\f(\val_j(\p)), \Traces_n(i), 0) \] 
The $(i,j)^{th}$ element of the matrix represents the robustness of
the $i^{th}$ trace, $\Traces_n(i)$ with respect to the $j^{th}$ \PSTL
formula $\f(\val_j(\p))$. For checking the satisfaction of a STL
specification by a trace we use Breach \cite{donze2010breach}, a
toolbox for verification and parameter synthesis of hybrid systems.
This procedure is implemented in {\em computeSignature} function in
Algorithm\ref{alg:logical_classification}. {\em computeSignature}
function in Algorithm\ref{alg:logical_classification} is used to
detect whether an enumerated PSTL formula $\f$ is new or its
equivalent has been enumerated. 


%
%
%


Using the idea of signatures reduced the computation time in targeted
case studies; the summary of the results are shown in Table
\ref{tab:signature}. More explanation about computing signature for
each case will be provided in Sec.~\ref{sec:casestudies}. As Table
\ref{tab:signature} shows, the computation time difference before and
after optimization is noticeable, yet, it is at best 22\%.  The reason
is that the enumerative solver produces simple \STL classifiers in 5
case studies, and until reaching those \STL classifiers, only a few
formulas with equivalent signatures are enumerated. For producing
complicated classifiers, the difference would be more noticeable.

\begin{table}[ht]
\centering
\begin{tabular*}{.7\textwidth}{@{\extracolsep{\fill}}ccc}
\toprule
Case & Before   & After  \\
&optimization (s)&Optimization (s)\\
\midrule
Linear system&44.25&39.05 \\
Cruise control of train&35.84&32.31 \\
Human Activity Recognition&28.06&22.07  \\
Robot execution failures & 29.64 & 27.91  \\
Environment assumption mining & 158.70 & 126.66  \\ [1ex]
\bottomrule
\end{tabular*}
\caption{Optimization using signature\label{tab:signature}}
\end{table}

\section{Logical Binary Classification}
\label{sec:logic_bin_classification}
Next, we explain how to learn a logical and interpretable classifier
for binary classification. We assume that we are given two sets of
traces $\Traces^0$ (those with label $0$) and $\Traces^1$ (those with
label $1$). The main steps in the algorithm are as follows:

The function $\isNew$ checks whether the PSTL formula $\varphi(\p)$
produced by Algo.~\ref{alg:formula_enum} is heuristically equivalent
to an existing formula. Internally, this is done by checking if the
signature of the new formula is identical to the signature of an
existing formula in the database of formulas, i.e. $\db$. In the next
step, algorithm tries to obtain a point on the satisfaction boundary
of the enumerated formula $\varphi(\p)$ that results in a formula that
serves as a good classifier. To explain this procedure further, we
first recall the algorithm to approximate the satisfaction boundary of
a given formula $\varphi(\p)$.

First we recall from \cite{vazquez2018time} that the satisfaction
boundary of a formula $\varphi(\p)$ with respect to a set of traces
$\Traces$ is essentially the set of parameter valuations $\val(\p)$
where the robustness value of $\varphi(\val(\p))$ for at least one
trace is $0$, i.e. the formula is marginally satisfied by at least one
trace in $\Traces$.

\begin{figure}[!t]
\centering
\includegraphics[scale=0.12]{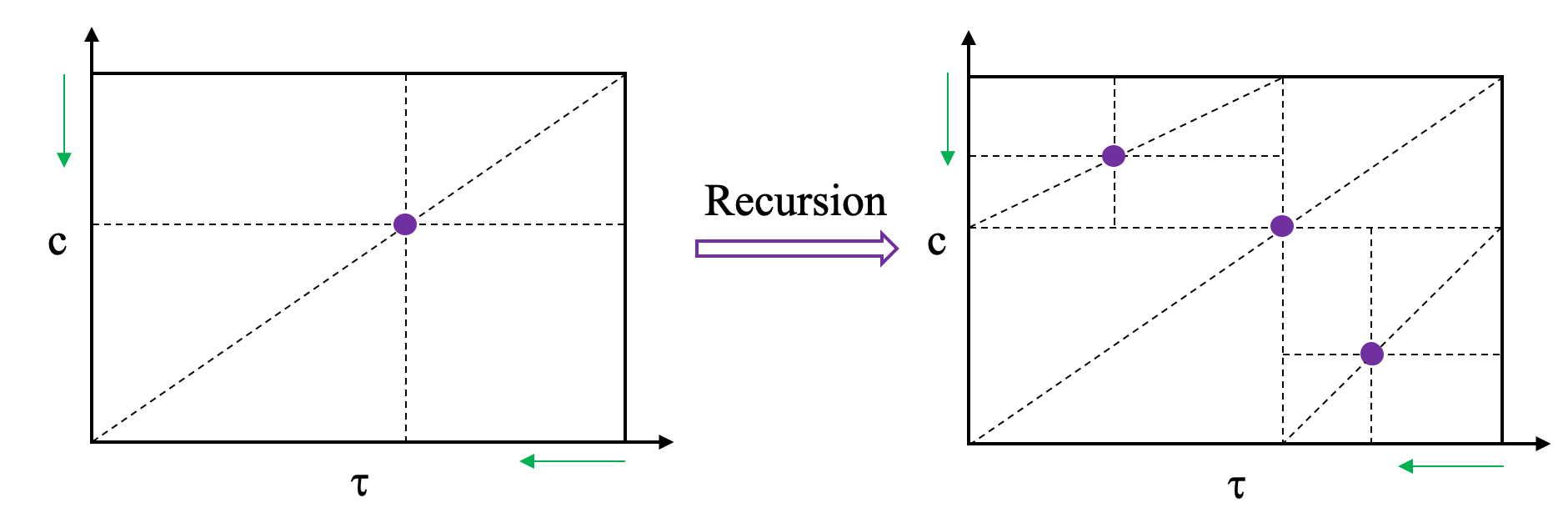}
\caption{Method to recursively approximate satisfaction boundary to
arbitrary precision $\delta$. Green arrows indicate the monotonicity
direction (both decreasing).}
\label{fig:mapping}
\end{figure}

In general, computing the satisfaction boundary for arbitrary PSTL
formulas is difficult;  however, for formulas in the monotonic
fragment of \PSTL \cite{vazquez2018time}, there is an efficient
procedure to approximate the satisfaction boundary \cite{maler2017learning}.  The
procedure in \cite{maler2017learning} recursively approximates the
satisfaction boundary to an arbitrary precision by performing binary
search on diagonals of sub-regions within the parameter space. The
idea is that in an $m$-dimensional parameter space, a parameter
valuation on the diagonal corresponds to a formula with zero
robustness value. This point subdivides the parameter space into $2^m$
distinct regions: one where all valuations correspond to formulas that
are valid over all traces, one where all valuations correspond to
formulas that are invalid, and $2^m - 2$ regions where satisfaction is
unknown. The algorithm then proceeds to search along the diagonals of
these remaining regions.  This approximation results in a series of
overlapping axis-aligned hyper-rectangles guaranteed to include the
satisfaction boundary \cite{vazquez2018time}. More details of this
procedure can be found in \cite{maler2017learning}.  We visualize an
instance of the method in Fig.~\ref{fig:mapping}.

We combine the procedure from \cite{maler2017learning} with a
classification algorithm as follows. In each recursive iteration of
the multi-dimensional binary search, the algorithm identifies a point
on the satisfaction boundary of $\f(\p)$ with respect to the
$1$-labeled traces, i.e.  $\Traces^1$. Let this point be denoted
$\val^*(\p)$, and the resulting STL formula $\f(\val^*(\p))$ be
denoted as $\f^*$ in short-hand.  We then check the Boolean
satisfaction of $\f^*$ on the traces in $\Traces^0$. The
mis-classification rate (\textit{MCR}) is computed as follows:
\begin{equation}
\mathit{MCR} = \frac{|\setof{x \mid x \in \Traces ^0 \wedge x \models \f^*}|}
                    {|\Traces^0| + |\Traces^1|}
\end{equation}

If \textit{MCR} is less than the specified threshold (threshold =
$0.1$ in our implementation), the algorithm terminates, and $\f^*$ is
returned as the binary \STL classifier for traces $\Traces^0$ and
$\Traces^1$. Otherwise, the algorithm goes to the next recursive
computation of a boundary point (in another region of the parameter
space). If the size of the parameter-space sub-region being searched
(denoted by the variable $boundaryPrecision$ in
Algo.~\ref{alg:logical_classification}) exceeds a user-specified
$\delta$, the algorithm returns control to
Algo.~\ref{alg:formula_enum}, which proceeds to enumerate the next
PSTL formula for consideration.

Algo.~\ref{alg:formula_enum} continues execution
until a \PSTL classifier $\varphi$ with
\textit{MCR $<$ threshold} is found or the length of enumerated \PSTL
formula exceeds the \textit{maxLength} defined by user, which means no
\PSTL classifier with \textit{MCR $<$ threshold} can be found.

\begin{algorithm}[!t]
\small
\KwIn{$\f, \Traces^0, \Traces^1, \delta, threshold, \db$}

\SetKwProg{Fn}{Function}{:}{}

\If{$isNew(\varphi)$}{
	\While{$boundaryPrecision < \delta$}{
	    \tcp{Selects a candidate point on satisfaction boundary}
	    $\val^*(\p) \leftarrow pointOnBoundary(\varphi, \Traces^1)$\;   
	    \tcp{Replaces candidate point on satisfaction boundary in $\varphi$}
	    $\f^* \leftarrow \f(\val^*(\p))$\;
	    \tcp{Compute Boolean satisfaction for traces with label 0}
		$falsePos \leftarrow \lvert\setof{x \mid x \in \Traces^0 \wedge x \not\models \f^*\rvert}$ \;
		$MCR \leftarrow falsePos / (\lvert T^0 \rvert +  \lvert T^1 \rvert)$\;
		\lIf{$MCR < threshold$}{ \Return $\varphi, MCR$\; }
	}
		
}
\vspace{0.05cm}
\Fn{$isNew()$}{
     $S_\f \leftarrow computeSignature(\varphi)$\;
     \tcp{Compare signature with formula signatures in the database}
     \lIf{$S_\varphi \in \db$}{
     	\Return $false$\;
     }
     \Else{
         \tcp{Add signature to database}
         $Add(\db, \varphi, S_\varphi)$\;
         \Return $true$\;
     }
}
\vspace{0.05cm}
\Fn{$computeSignature()$}{
	\tcp{Randomly choose $n$ traces}
	$\Traces_n \leftarrow selectRandom(\Traces^0, \Traces^1, n)$\;
	\tcp{Randomly sample $m$ parameter values}
	$\paramspace_m \leftarrow sampleRandom(\paramspace,m)$\;
	\For{$j \leftarrow 1$ to $m$}{
		$\varphi_m \leftarrow setParams(\varphi_m, \paramspace_m(j))$\;
		\For{$i\gets 1, n$}{
			$S_\varphi (i,j) \leftarrow \rho(\varphi_m, \Traces_n(i), 0)$\;

		}
	}
}
\caption{Logical binary classification \label{alg:logical_classification}}
\end{algorithm}

\section{Case Studies}
\label{sec:casestudies}
To evaluate the new framework, we apply our method on various case
studies from the automotive, transportation and healthcare
domain\footnote{We run the experiments on an Intel Core-i7 Macbook Pro
with 2.7 GHz processors and 16 GB RAM.}. The results show that the
employed technique has a number of advantages compared to previous
methods which are mentioned in section \ref{sec:conc}.

\begin{figure}[!t]
\centering
\includegraphics[scale=0.15]{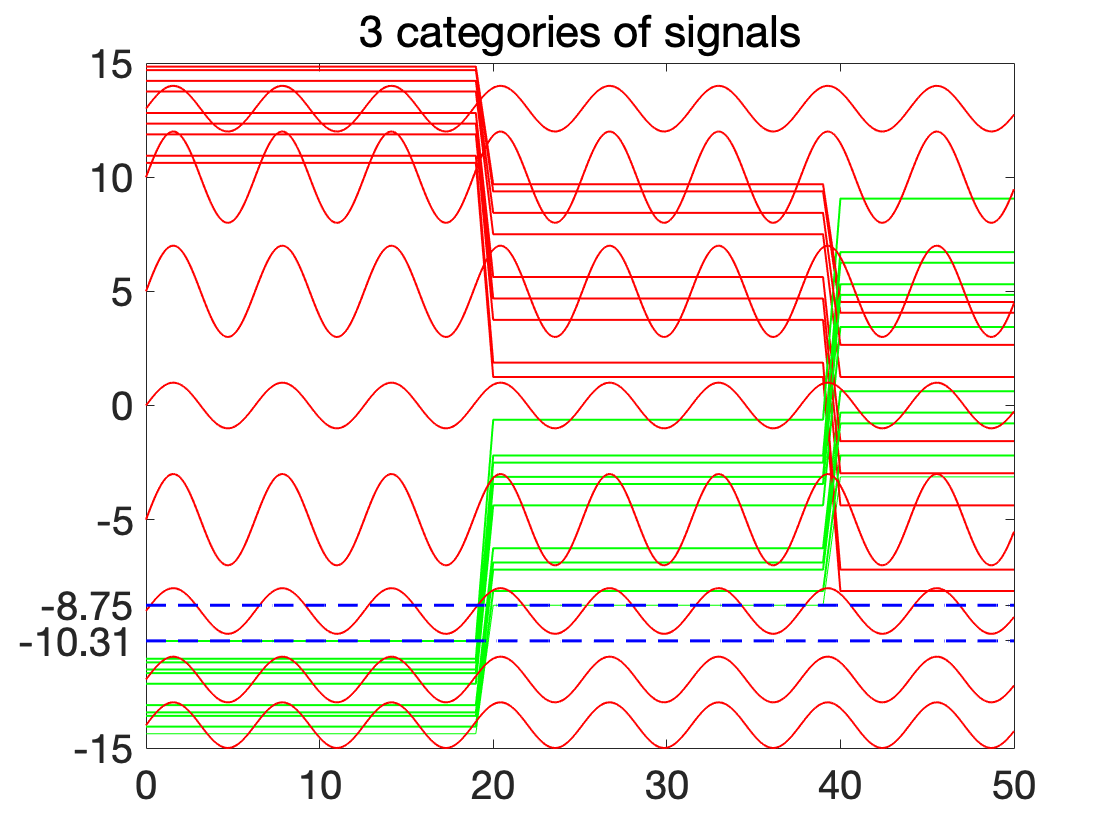}
\caption{Synthetic traces (green traces: upward steps, red traces: downward steps and
sinusoids and dash lines: the thresholds of \STL learned by
enumerative solver ($= -10.31$ and $-8.75$)).}
\label{fig:synthetic_data}
\end{figure}

\mypara{Experiments on Synthetic data} We generated 28 synthetic
traces (10 upward steps, 10 downward steps, and 8 sinusoids) in
Fig.~\ref{fig:synthetic_data} in green and red colors. We are
interested in categorizing green traces (upward steps) from red ones
(downward steps and sinusoids). We applied our enumerative solver on
these traces, and the resulting \STL classifier is: $(x[t] < -10.31)
\aand \ev_{[0,39.58]}(x[t] > -8.75)$, with $MCR = 0$ and execution
time = 4416.71 seconds.  This \STL classifier is compatible with the
general behavior of upward steps. The blue dash lines in
Fig.~\ref{fig:synthetic_data} show the thresholds of the learned \STL
formula ($-10.31$ and $-8.75$). In this case study, a signature with
$n=3$ and $m=5$ was used to detect the equivalent \PSTL formulas.  Of
the 48 formulas enumerated, there was only one equivalent formula
pruned by the signature-based equivalence check.  This case study
seeks to highlight an STL formula with a nested temporal operator. 

\begin{figure}[!t]
\centering
\includegraphics[scale=0.2]{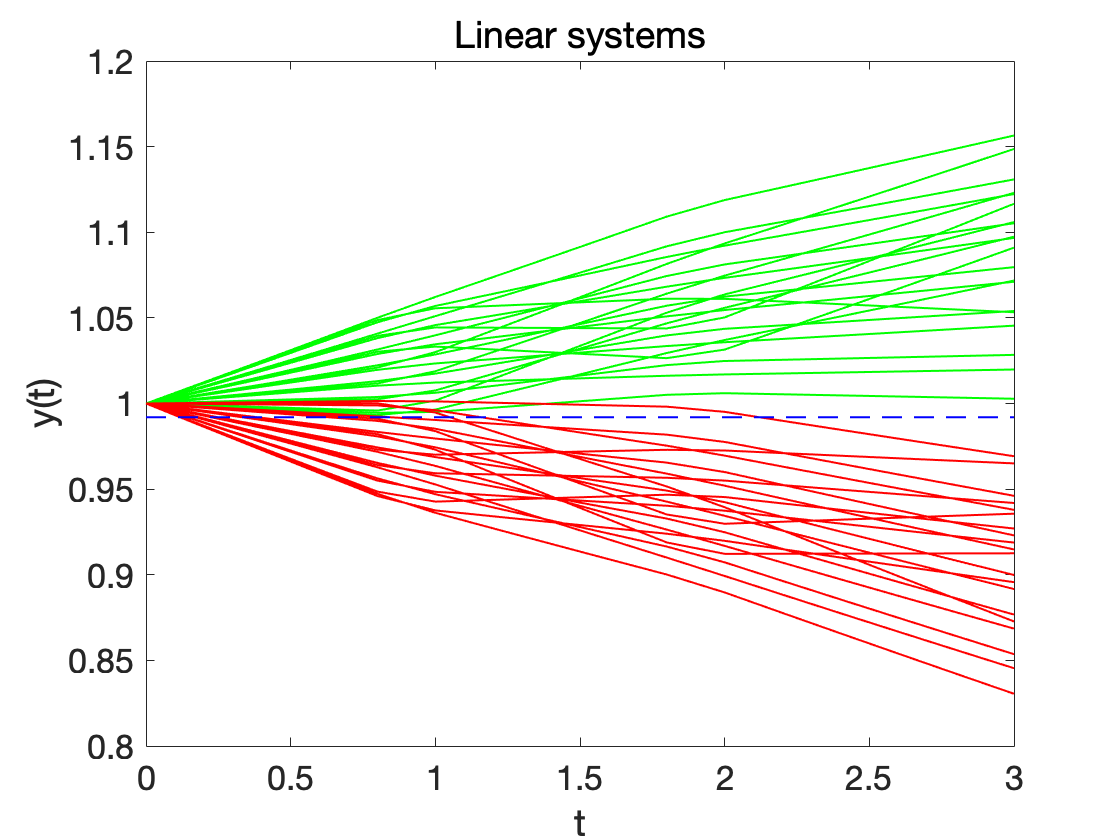}
\caption{Simulation results of the linear system (Green traces: normal operation of the system, red traces: anomalous behavior of the system and dash line: the threshold of the \STL learned by enumerative solver (= 0.992)).}
\label{fig:belta_linear_system}
\end{figure}

\mypara{Linear System} We now compare our technique with another
supervised learning technique, and for a fair comparison, we use the
same models as \cite{jones2014anomaly}. We modeled the system $S(t)$
whose dynamics evolve according to Eq.~\eqref{eqn:belta_lin_sys} shown
in the Appendix.  We generated 100 time-series traces of the system
for the two different system modes, resulting in a total of 200
time-series traces.  Fig.~\ref{fig:belta_linear_system} shows the
results of simulation.  Green traces represent normal behaviors
(absence of attack), and red traces represent the behavior of the
system under attack. 50$\%$ of the data was split for training (50
normal and 50 anomaly), and the remaining data was reserved for
testing.
The enumerative solver was trained and tested on this data to extract
an \STL formula. The dashed blue line in the figure shows the threshold
(=0.992) of the learned formula: $\G_{[0,3]}(x(t) \geq 0.992)$. This is a simpler formula compared
to the one obtained in \cite{jones2014anomaly}, which is
$\F_{[0,3.0)}(\G_{[0.5,2.0)}(y > 0.9634))$.

We use a marginally better computing platform than Jones et al., and
our procedure takes 39.05s (with signature-based pruning) and 44.25s
(without signatures), whereas the approach by Jones et al. required
approximately 130s to extract the formula.  We used a signature with
$n=3$ and $m=5$ to detect equivalent \PSTL formulas. In total, 5
formulas were enumerated, and we could prune 2 of them as logically
equivalent to other enumerated formulas.

\mypara{Cruise Control of Train} We also benchmark an example for
cruise control in a train from \cite{jones2014anomaly}. The system
consists of a 3-car train, where each car has its own
electronically-controlled pneumatic (ECP) braking mechanism. The
velocity of the entire train is modeled as a single system. Hence,
there are a total of 4 models (1 for velocity + 3 for braking in each
car). The description of the velocity and ECP braking models can be
found in the Appendix. The observations/readings
of the train velocity is assumed to be noisy.

\begin{figure}[!t]
\centering
\includegraphics[scale=0.2]{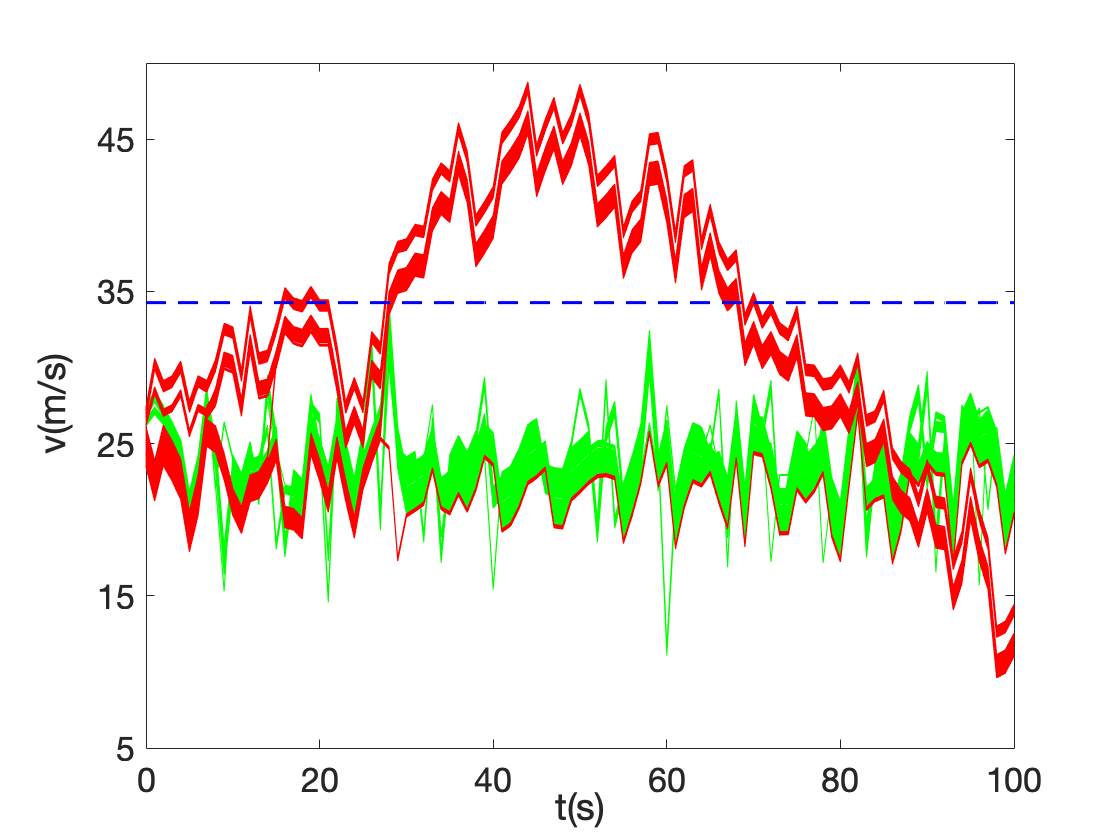}
\caption{Simulation results of cruise control of the train (Green traces: normal operation, red traces: anomalous behavior and dash line: the threshold of the \STL learned by enumerative solver (= 34.2579)).}
\label{fig:train_cruise_traces}
\end{figure}

The cruising speed of the train is set to $25m/s$ and the train
oscillates about this speed by $\pm 2.5m/s$. Under normal conditions,
the train maintains the cruising speed and applies its brakes when the
speed exceeds a threshold/upper limit. In an anomalous situation (or
attack), all the brakes fail to engage and hence the train fails to
maintain its speed within the desired/set limits.  The velocity
parameters were set to be in the interval $[0, 30] m/s$. For this
system, we generated data from 200 simulations (100 for normal and 100
for anomaly behavior). The time-series data corresponding to the
simulations is shown in Fig.~\ref{fig:train_cruise_traces}, in which
the green traces represent normal behaviors, while red traces
represent anomalies. Since Breach \cite{donze2010breach} could
possibly choose a low initial velocity during an anomaly situation, a
very low percentage of the traces tend to exhibit normal behaviors, as
seen from the figure. Similar to our previous approaches, the data was
split 50$\%$ (50 normal and 50 anomaly traces) for training and the
rest for testing. 

We applied our solver to extract the STL formula: $\G_{[0,100]}(x(t) >
34.2579)$, where the threshold learned by our solver is 34.2579 (shown
by the dashed blue line in the figure). The \textit{MCR}s for training
and testing are $0.05$ and $0.04$ respectively. The \STL formula
obtained by our approach is simpler compared to the one extracted in
\cite{jones2014anomaly}, which is: $\F_{[0,100)}( \F_{[10,69)}(y <
24.9) \aand \F_{[13.9,44.2)}(y > 17.66)) )$.  The total time for
learning the classifier in \cite{jones2014anomaly} 154s, while the
execution time of our approach is 32.31s (with signatures) and 35.84s
(without signatures). In this case study, a signature with $n=3$ and
$m=5$ was used to detect the equivalent \PSTL formulas. Of the 5
enumerated formulas, we used signatures to prune 2 formulas. 

\mypara{Human Activity Recognition}

The UC Irvine time-series learning repository contains time-series
data for Human Activity Recognition from inertial sensors (such as
accelerometers) in waist-mounted smartphones \cite{anguita2013public}.
This data was obtained from the recordings of 30 subjects performing
activities of daily living.  Six basic activities were performed by
human subjects: three static postures (standing, sitting, lying) and
three dynamic activities (walking, walking downstairs and walking
upstairs).  In our experiments, we consider the classification of the
data into the two postures: static and dynamic. This is an important
task for healthcare professionals to monitor the activities and vital
organs of people undergoing medication and the elderly. The
time-series signals from the accelerometers are processed using
certain filters. One of them computes the standard deviation of the
acceleration in the $x$ direction in a certain time-window. We used
this signal to study whether we can obtain STL classifiers.

\begin{figure}[!t]
\centering
\includegraphics[scale=0.2]{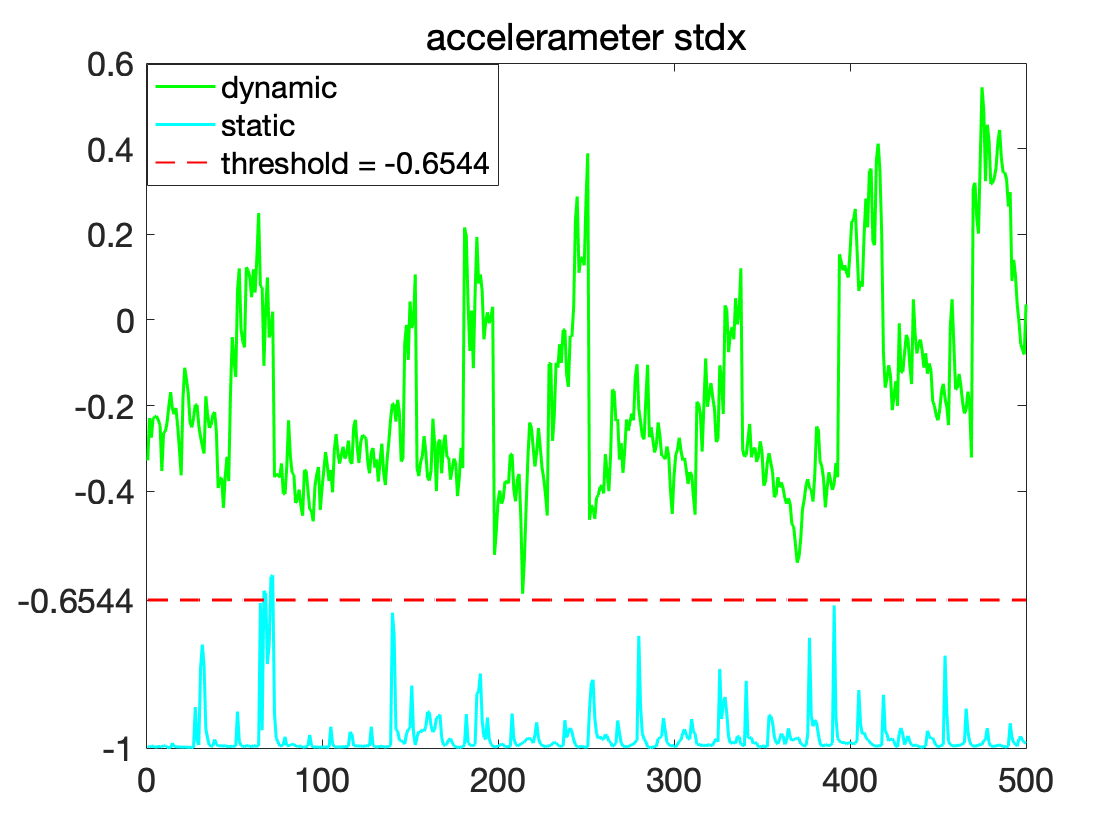}
\caption{Human activity recognition (green trace: dynamic, blue trace: static, and dash line: the threshold of \STL learned by
enumerative solver (= -0.6544)).}
\label{fig:har}
\end{figure}

A sampling of the corresponding data is shown in Fig.~\ref{fig:har}.
25 traces, each having a sampling rate of 50 samples per second for a
total of 10 seconds, were used for training. 10 traces were used for
testing. We were able to obtain an \textit{MCR} of $0$ on both the
training and test data. The \STL formula learned by the algorithm is:
$\G_{[0,49]} (stdx[t] > -0.6544)$, where $stdx$ is the aforementioned
standard deviation signal provided by the data.  Current methods
\cite{Anguita2013EnergyES, har_svm_davide, har_ortiz_luis2014} achieve
an accuracy of about 96\%. Our solver was able to achieve an accuracy
of 100\%, which shows a significant improvement over current
approaches \cite{Anguita2013EnergyES, har_svm_davide,
har_ortiz_luis2014}. Furthermore, our solver uses only 1 feature
compared to \cite{Anguita2013EnergyES}, which uses 17 features.  The
time required for execution of the solver is 22.07 seconds (with
signatures) and 28.06 seconds (without signatures).

While we find our results a significant improvement over the
state-of-the-art in terms of accuracy and run-time, we comment that ML
algorithms excel at being able to learn from large feature spaces, and
choosing the right feature which is a challenging task. It would be
interesting to benchmark our approach where we use all the signal
variables in the data (and the possibly much larger space of formulas
over the larger number of signals).

\mypara{Robot Execution Failures}\label{ssec:robot_failure}

The robot execution failures dataset contains force and torque
measurements along the 3 axes (x, y and z) of an assembly
(pick-and-place) robot after detecting a failure
\cite{anguita2013public}. $15$ force/torque samples collected at
regular time intervals characterize each failure.  It consists of 5
datasets of failures in the following tasks: approach to grasp
position, transfer of a part, position of part after a transfer
failure, approach to ungrasp position and in motion with part. The
dataset is obtained from the UCI Machine Learning repository at
\cite{Dua:2019}. In our experiments, we use the two of the datasets of
robot failures, which are described as follows.

\noindent{\em Dataset LP2: Failures in transfer of a part.} This
dataset consists of 88 traces and contains classes representing
normal, collision, front collision and obstruction behaviors. For our
experiment, we used the 21 traces that exhibit normal behavior and 17
traces that exhibit collision, and applied our solver to classify the
traces into these two groups. The results of our solver are shown in
Table \ref{tab:lp2_fail}. In \cite{lopes1998feature}, their best
algorithm had an accuracy of 68\% which uses \textit{Fourier
transform} of force and torque in three direction of x, y, and z. Our
solver shows a major improvement in accuracy as the misclassification
rate is less than 0.1\%, by using only one feature from raw data. The
result \STL classifiers using each of the three features $F_x$, $T_x$,
and $T_y$  are shown in Table \ref{tab:lp2_fail}.
\begin{table}[t]
\centering
\scalebox{0.8}{
\begin{tabular}{ccccc}
\toprule
Feature & STL classifier  & Time before & Time after & MCR \\
&  & optimization (s) & optimization (s) &  \\
\midrule
$F_x$ & $\G_{[0,14]} (F_x [t] < 2)$ & 29.65 & 27.92 & 0.026 \\
$T_x$ & $\G_{[0,14]} (T_x [t] > -14)$ & 26.69 & 26.54 & 0.078 \\
$T_y$ & $\G_{[0,14]} (T_y [t] > -9.0002)$ & 30.95 & 28.64 & 0.078 \\
\bottomrule
\end{tabular}}
\caption{Robot execution failure \label{tab:lp2_fail}}
\end{table}

\begin{figure}[!t]
\centering
\includegraphics[scale=0.2]{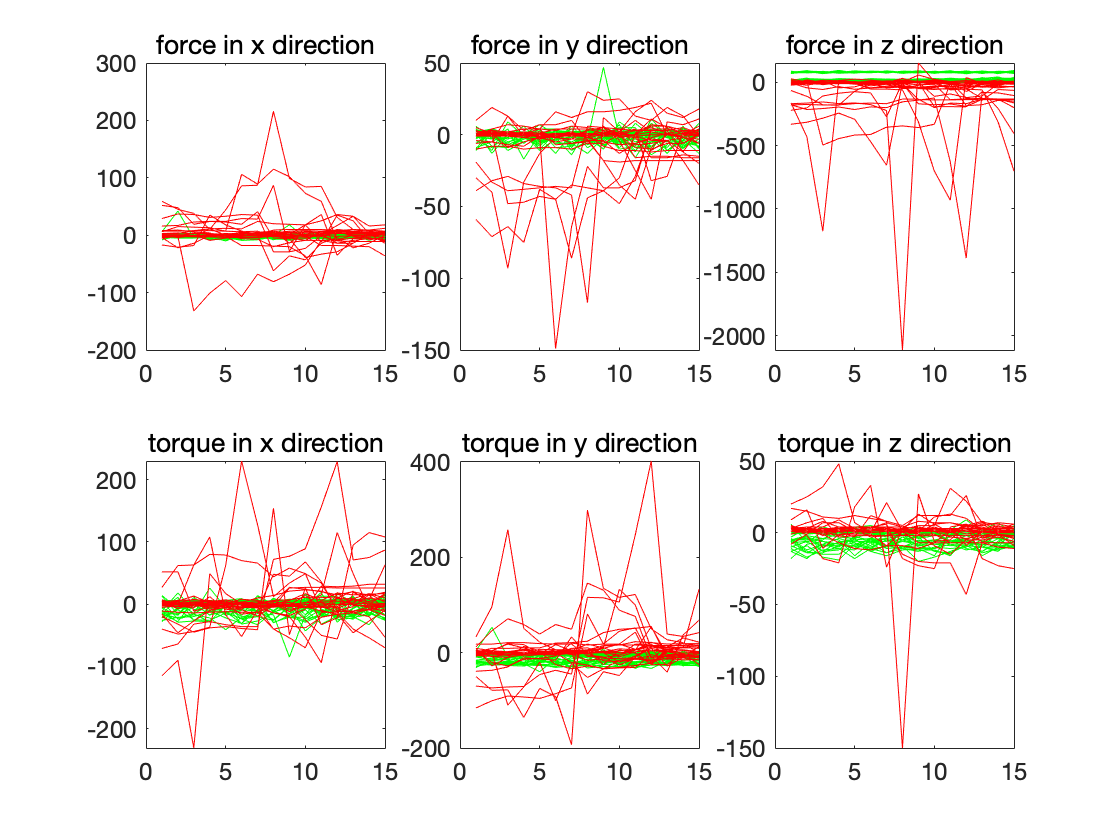}
\caption{Robot Failure Execution (Green traces: normal behaviors, red traces: collisions).}
\label{fig:robot_failure_lp5}
\end{figure}

\noindent{\em Dataset LP5: Failures in motion with part.} This dataset
consists of 164 instances, of which 44 instances exhibit normal
behavior and 26 instances exhibit bottom collision. The dataset
containing these traces are shown in Fig.~\ref{fig:robot_failure_lp5} 
in which the green traces indicate normal
behaviors, while the red traces indicate collisions. We used the same
approach as for the LP2 dataset and our solver learned the \STL
$\G_{[0,4]} (F_z[t] \geq 0)$, with a misclassification rate of 0. The
processing time with signature is $22.83 s$, while the processing time
without signature is $22.63 s$. In this case, the processing time
without using signature is better; the reason is that until reaching
the ``always'' temporal logic operator, we only encounter two
repetitive formulas which are atomic predicates $(\neg x[t] < valx)$
and $(\neg x[t] > valx)$ for which, the processing times are very
small, while the times required to compute the signature are greater.
Therefore, signature is more effective when the length of the formula
and number of traces increase. In \cite{lopes1998feature}, their best
algorithm had an accuracy of 77\% which uses a long feature vector.
Our solver, with 0\% misclassification rate by using only 1 feature
$F_x$ shows a remarkable improvement compared to
\cite{lopes1998feature}.

\begin{figure}[!t]
\centering
\includegraphics[scale=0.2]{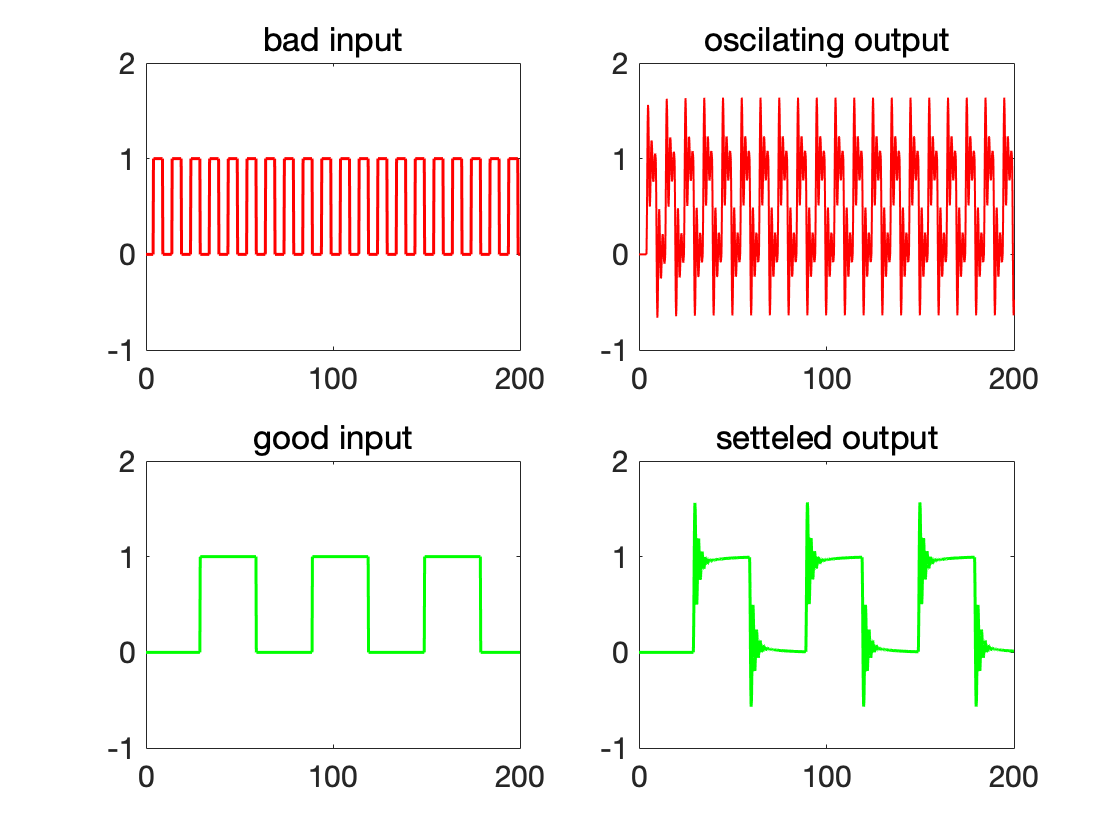}
\caption{Environment assumption mining (red: bad input leads to undesirable output, green: good input leads to desirable output).}
\label{fig:env_assumption_mining}
\end{figure}

\mypara{Environment Assumption Mining} Given a deterministic system
$S(t)$ and a specification $\phi$, the goal of assumption mining is to
determine the subset of initial states or inputs that ensure
satisfaction of an STL formula $\varphi$ on the output of the model.
In other words, we want to know what \STL formulas do the inputs
satisfy that guarantee that the outputs will be desirable.  In our
experiments, we try to learn the \STL formulas that separates good
inputs from bad inputs.  For our experiment, we consider a PID
controller for a damped second-order continuous system. The desired
output of this system is to observe that the output has settled or is
oscillating.  We use 31 input traces with periods: 10, 12, 14, ..., 70
and their negations.  The traces are shown in
Fig.~\ref{fig:env_assumption_mining}. For oscillating outputs, the
\STL formula learned by the solver is: $\G_{[0,368]} (\F_{[0,21]}
(x[t] \geq 1))$. It means for periods below $21 \times 2 = 42$, the
output does not settle. This is to be expected, as this period is less
than the settling time of the system.
The time required for learning is 126.66s (with signatures)
and is 158.70s (without signatures). In this case, $n=2$ and $m=2$
were chosen as dimensions of signature. Among total of 11 formulas
enumerated, signatures could prune 4 equivalent formulas.

\section{Related Works}
\label{sec:conc}
There has been considerable recent work on learning STL formulas from
data for various applications such as supervised learning
\cite{bombara2016decision,kong2014temporal}, clustering
\cite{vazquez2017logical,vazquez2018time}, or anomaly
detection\cite{jones2014anomaly}.

In \cite{kong2014temporal}, a fragment of PSTL (rPSTL or reactive
parametric signal temporal logic) is defined to capture causal
relationships from data. However, there are some temporal properties
namely, {\em concurrent eventuality} and {\em nested always
eventually} that cannot be described directly in rPSTL.  In
\cite{jones2014anomaly}, the authors extend \cite{kong2014temporal} by
using a fragment of rPSTL, inference parametric STL (iPSTL), that does
not require a causal structure. In this work, classical ML algorithms
(one-class support vector machines) are applied for unsupervised
learning problem. In \cite{bombara2016decision}, a decision tree based
method is employed to learn STL formulas, which creates a map between
a restricted fragment of STL and a binary decision tree in order to
build a STL classifier.  While this seminal work has advanced work in
the intersection of formal methods and machine learning, one
disadvantage of these approaches has been that they lead to long
formulas which can become an issue for interpretability.

In template-based techniques, a fixed PSTL template is provided by the
user, and the techniques only learn the values of parameters
associated with the PSTL. In \cite{vazquez2017logical}, a total
ordering on parameter space of PSTL specifications is utilized as
feature vectors for learning logical specifications. Unfortunately,
recognizing the best total ordering is not straightforward for users.
In \cite{vazquez2018time}, the authors eliminate this additional
burden on the user by suggesting a method that maps timed traces to a
surface in the parameter space of the formula, and then employing
these curves as features. In \cite{jin2015mining}, the input to the
algorithm is a requirement template expressed in PSTL, where the
traces are actively generated from a model of the system.  Our
proposed technique, which uses systematic enumeration, can produce
smaller formulas which may be more human-interpretable, and with
higher accuracy($\geq 92\%$ in all investigated case studies).

\section{Conclusion}
We proposed a new technique for binary classification of time-series
data using Signal Temporal Logic formulas.  The key idea is to combine
an algorithm for systematic enumeration of PSTL formulas with an
algorithm for estimation of the satisfaction boundary of the
enumerated PSTL formula.  We also investigate an optimization using formula signatures to avoid enumerating equivalent PSTL
formulas. We then illustrate this technique with a number of case
studies on real world data from different domains. The results show
that the enumerative solver has a number of advantages.  As future
work, we will extend this approach to multi-class classification,
unsupervised, semi-supervised, and active learning. We will also
investigate other optimization techniques to make the enumerative
solver faster and more memory-efficient.

\bibliographystyle{unsrt}  
\bibliography{references}  

\end{document}